\documentclass[runningheads]{llncs}
\usepackage{graphicx}
\usepackage{style}
\usepackage{amsfonts} 
\usepackage{cite}
\usepackage{amsmath}
\begin{document}
\title{Raw or Cooked? \\Object Detection on RAW Images}
%
%

\author{William Ljungbergh\inst{1,2}\orcidID{0000-0002-0194-6346} \and
Joakim Johnander\inst{1,2}\orcidID{0000-0003-2553-3367} \and
Christoffer Petersson\inst{2}\orcidID{0000-0002-9203-558X} \and 
Michael Felsberg\inst{1}\orcidID{0000-0002-6096-3648}
}

\authorrunning{W. Ljungbergh et al.}


\institute{Computer Vision Laboratory, Linköping University, 581 83 Linköping, Sweden \email{\{william.ljungbergh, michael.felsberg\}@liu.se} \and
Zenseact, Lindholmspiren 2, 417 56 Gothenburg, Sweden \\
\email{\{joakim.johnander, christoffer.petersson\}@zenseact.com}\\
}
\maketitle              

\begin{abstract}
Images fed to a deep neural network have in general undergone several handcrafted image signal processing (ISP) operations, all of which have been optimized to produce visually pleasing images. In this work, we investigate the hypothesis that the intermediate representation of visually pleasing images is sub-optimal for downstream computer vision tasks compared to the RAW image representation. We suggest that the operations of the ISP instead should be optimized towards the end task, by learning the parameters of the operations jointly during training. We extend previous works on this topic and propose a new learnable operation that enables an object detector to achieve superior performance when compared to both previous works and traditional RGB images. In experiments on the open PASCALRAW dataset, we empirically confirm our hypothesis.

\end{abstract}
\section{Introduction}
Image sensors commonly collect RAW data in a one-channel Bayer pattern~\cite{bayer1976color,langseth2014evaluation}, \emph{RAW images}, that are converted into three-channel RGB images via a camera Image Signal Processing (ISP) pipeline. This pipeline comprises a number of low-level vision functions -- such as decompanding~\cite{hp2020autoencoder}, demosaicing~\cite{hirakawa2005adaptive} (or \emph{debayering}~\cite{langseth2014evaluation}), denoising, white balancing, and tone-mapping~\cite{suma2016evaluation,mujtaba2022efficient}. Each function is designed to tackle some particular phenomenon and the final pipeline is aimed at producing a visually pleasing image.

In recent years, image-based computer vision tasks have seen a leap in performance due to the advent of neural networks. Most computer vision tasks -- such as image classification or object detection -- are based on RGB image inputs. However, some recent works~\cite{omid2014pascalraw,zhang2021raw} have considered the possibility of removing the camera ISP and instead directly feeding the RAW image into the neural network. The intuition is that the high flexibility of the neural network should enable it to approximate the camera ISP if that is the optimal way to transform the RAW data. It is important to note that the camera ISP is in general not optimized for the downstream task, and the neural network might by itself be able to learn a more suitable transformation of the RAW data during the training. One possibility is that the ISP might remove information that could be crucial in adverse conditions, such as low light. Moreover, the camera ISP adds image data according to image priors, which might result in spurious network responses~\cite{kriesel2014traue}. 

\begin{figure*}[t]
    \centering
    \begin{tabular}{ccc}
        \includegraphics[width=0.32\linewidth]{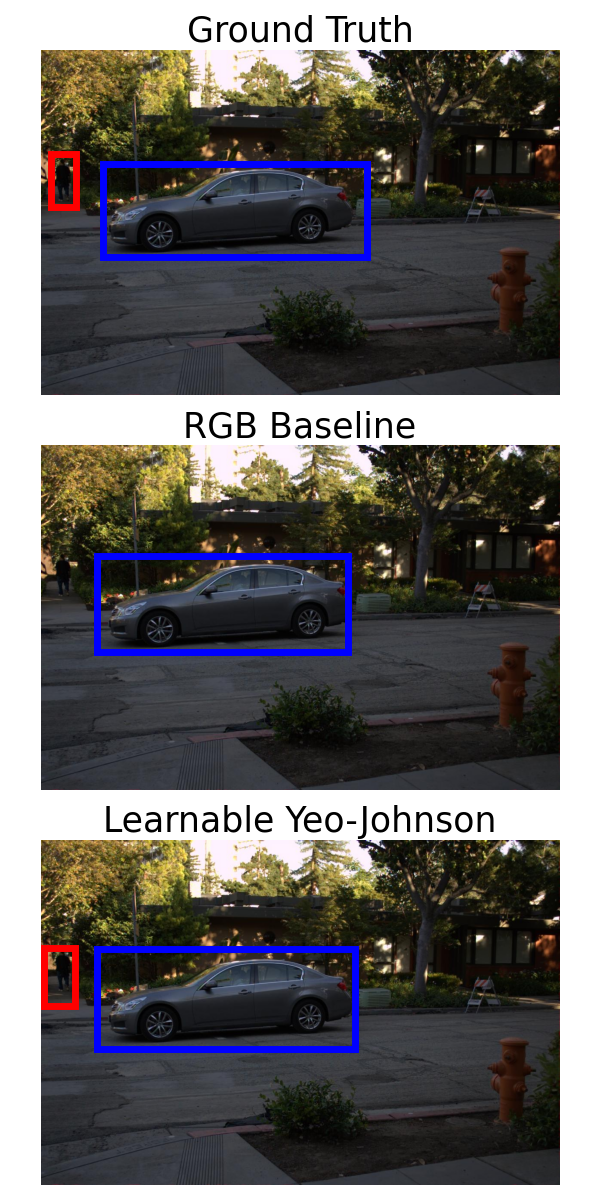} &
        \includegraphics[width=0.32\linewidth]{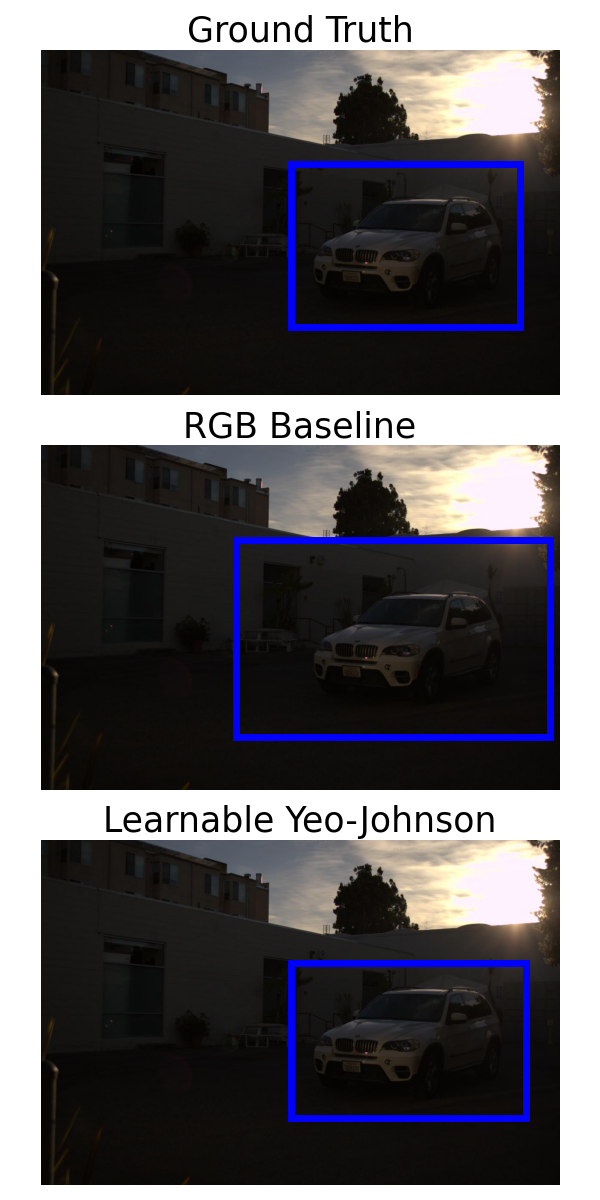} &
        \includegraphics[width=0.32\linewidth]{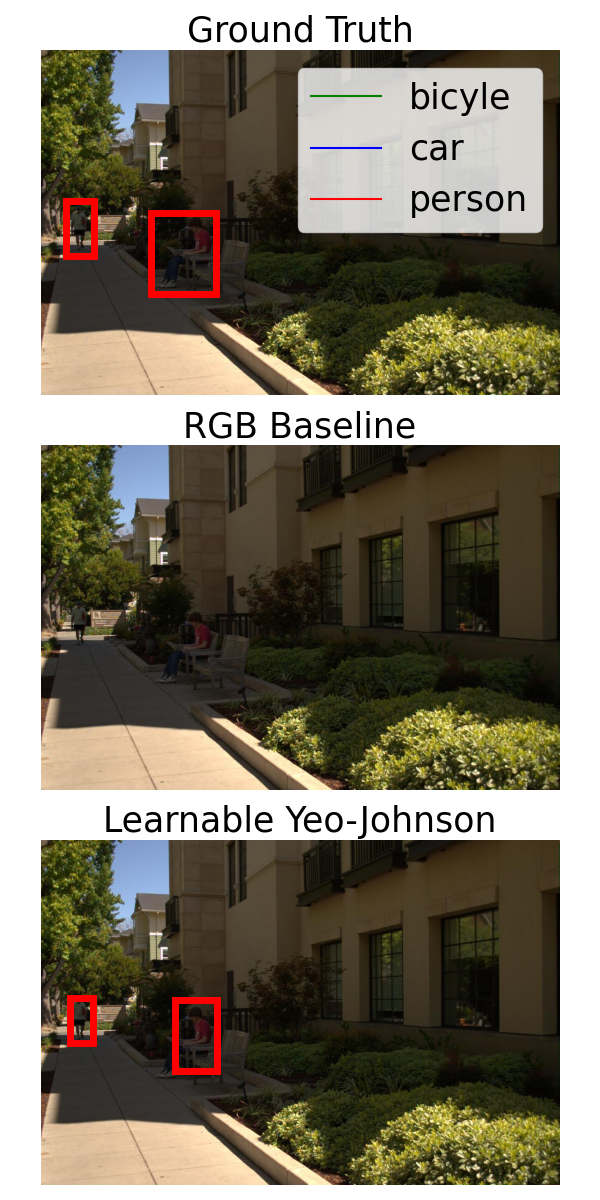} 
    \end{tabular}
    \caption{Three qualitative examples from the PASCALRAW dataset. We show the ground-truth (top), the RGB baseline detector (center), and the RAW RGGB detector with a learnable Yeo-Johnson operation (bottom). Compared to the RGB baseline, our proposed RAW RGGB detector manages to detect objects subject to poor light conditions.}
    \label{fig:samples-raw-better}
\end{figure*}

In this work we investigate object detection on RAW data, following the hypothesis that RAW input images lead to superior detection performance, with the aim to identify the minimal set of operations on the RAW data that results in performance that exceeds the traditional RGB detectors. Our main contributions are the following: 
\begin{enumerate}
\item We show that naïvely feeding RAW data into an object detector leads to poor performance.
\item We propose three simple yet effective strategies to mitigate the performance drop. The outputs of the best performing strategy -- a learnable version of the Yeo-Johnson transformation -- are visualized in Figure~\ref{fig:samples-raw-better}.
\item We provide an empirical study on the publicly available PASCALRAW dataset.
\end{enumerate}

\section{Related Work}
\parsection{Object detection} Object detection has been an active area of research for many years, and has been approached in many different ways. It is common to divide object detectors into two categories: (i) two-stage methods \cite{girshick2014rich, ren2015faster, lin2017feature} that first generate proposals and then localize and classify objects each proposal; and (ii) one-stage detectors that either make use of a predefined set of anchors \cite{lin2017focal, redmon2018yolov3} or make a dense (anchor-free) \cite{zhou2019objects, tian2019fcos} prediction across the entire image. Carion \etal~\cite{carion2020end} observed that both these categories of detectors rely on hand-crafted post-processing steps, such as non-maximum suppression, and proposed an end-to-end trainable object detector, DETR, that directly outputs a set of objects. One drawback of DETR is that convergence is slow and several follow-up works~\cite{zhu2020deformable, liu2021swin, zhang2022dino, wang2022anchor, meng2021conditional, sun2021rethinking} have proposed schemes to alleviate this issue. All the work above shares one property: they rely on RGB image data.

\parsection{RAW image data} RAW image data is traditionally fed through a \emph{camera ISP} that produces an RGB image. Substantial research efforts have been devoted into the design of this ISP, usually with the aim to produce visually pleasing RGB images. A large number of works have studied the different sub-tasks, \eg, demosaicing~\cite{malvar2004high,hirakawa2005adaptive,dubois2006filter,li2008image}, denoising~\cite{buades2005non,foi2008practical,condat2010simple}, and tone mapping~\cite{poynton2012digital,reinhard2002photographic,krawczyk2005lightness}. Several recent works propose to replace the camera ISP with deep neural networks~\cite{ignatov2020replacing,dai2020awnet,zhang2021learning,shekhar2022transform}. More precisely, these works aim to find a mapping between RAW images and high-quality RGB images produced by a digital single-lens reflex camera (DSLR).

\parsection{Object detection using RAW image data} In this work, we aim to train an object detector that takes RAW images as input. We are not the first to explore this direction. Buckler \etal~\cite{buckler2017reconfiguring} found that for processing RAW data, only demosaicing and gamma correction are crucial operations. In contrast to their work, we find that also these two can be avoided. Yoshimura \etal~\cite{yoshimura2022dynamicisp}, Yoshimura \etal~\cite{yoshimura2022rawgment}, and Morawski \etal~\cite{morawski2022genisp} strive to construct a learnable ISP that, together with an object detector, is trained for the object detection task. Based on our experiments, we argue that also the learnable ISP can be replaced with very simple operations. Most closely related to our work is the work of Hong \etal~\cite{hong2021crafting}, which proposes to only demosaic RAW images before feeding them into an object detector. In contrast to their work, we do not find the need for an auxiliary image construction loss nor for demosaicing.

\section{Method}
In this section, we first introduce a strategy for downsampling RAW Bayer images (Section \ref{section:downsampling}). This enables us to downsample high-resolution images to be more suitable for standard computer vision pipelines while maintaining the Bayer pattern in the RAW image. In Section \ref{sec:learnableisp}, we introduce the three \textit{learnable} operations.

\subsection{Downsampling RAW Images}
\label{section:downsampling}
\begin{figure*}[t]
    \centering
    \includegraphics[width=0.8\linewidth]{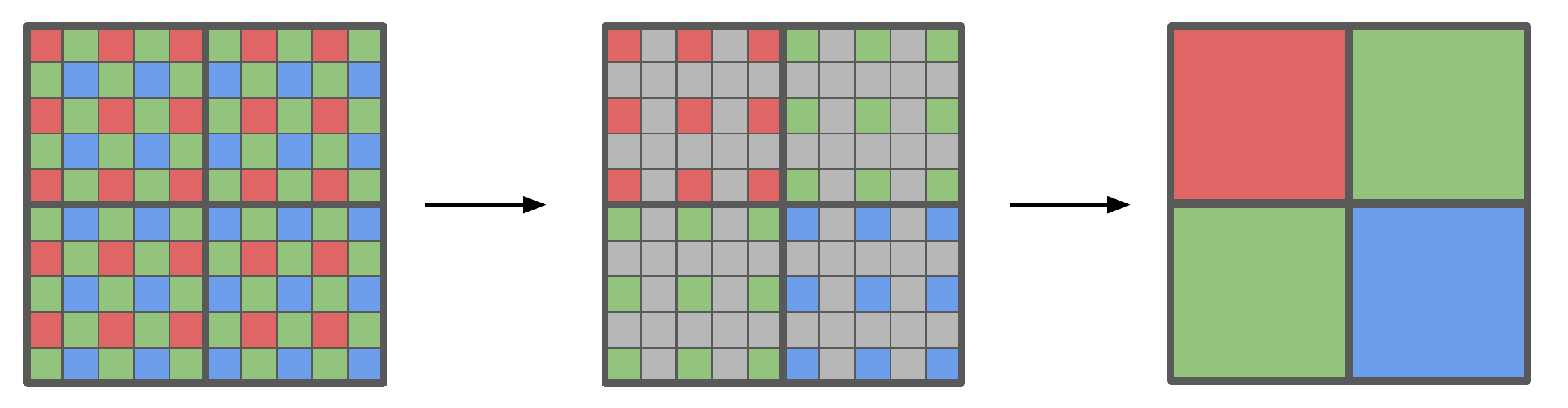}
    \caption{Downsampling method for Bayer-pattern RAW data. Each of the colors in the filter array of the downsampled RAW image (right) is the average over all cells in the corresponding region in the original image with the same color (left and center). The figure illustrates the downsampling of an original image patch of size $2d\times2d$ (with $d=5$ in this example), down to a patch of size $2\times2$, i.e. with a downsampling factor $d$ in each dimension.}
    \label{fig:bayer-downsample}
\end{figure*}

When working with high-resolution images, it is sometimes necessary to downsample the images to make them compatible with existing computer vision pipelines. However, standard downsampling schemes, such as bilinear or nearest neighbor, do not preserve the Bayer pattern that was present in the original image. To remedy this, we adopt a simple Bayer-pattern-preserving downsampling method, shown in Figure \ref{fig:bayer-downsample}.  Given an original RAW image $\mathbf{x}^\orig \in \mathbb{R}^{H \times W}$ and an uneven downsampling factor  $d \in 2\mathbb{N}+1$, we divide our original image into patches $x^\orig \in \mathbb{R}^{2d\times2d}$ with a stride $s = 2d$. Each patch is then downsampled by a factor $d$ in each dimension, yielding a downsampled patch $x \in \mathbb{R}^{2\times2}$, by averaging over the elements with the correct color in that sub-array. To clarify, all elements that correspond to a red filter in the upper left sub-array of the patch $x^\orig$ are averaged to produce the red output element $x_{0,0}$. The downsampling operation over the entire patch $x^\orig$ can be described as
\begin{equation}
    x_{i,j} = \frac{1}{N} \sum_{m =0}^{(d-1)/2} \sum_{n = 0}^{(d-1)/2} x^\orig_{di+2m, dj+2n} \enspace ,  
\end{equation}
where $x \in \mathbb{R}^{2\times2}$ is the downsampled patch,  $x^\orig\in \mathbb{R}^{2d\times2d}$ is the original patch, $d$ is the downsampling factor, $N = (d+1)^2/4$ is the number of elements averaged over, and $i,j \in 0, 1$. All downsampled patches are then concatenated to form the downsampled RAW image $\mathbf{x} \in \mathbb{R}^{H/d \times W/d}$.

It would be possible to feed the downsampled RAW image, $\textbf{x}$, directly into an object detector. There is however one thing to note about the first layer of the image encoder. In the standard RGB image setting, each weight in this layer is only applied to one modality -- red, green, or blue. This enables the first layer to capture color-specific information, such as gradients from one color to another. When fed with RAW images, as described above, we can assert the same property by ensuring that the stride of the first layer is an even number. Luckily, this is the case with the standard ResNet~\cite{he2016deep} architecture.

\subsection{Learnable ISP Operations}
\label{sec:learnableisp}
A standard ISP pipeline usually consists of a large collection of handcrafted operations. These operations are in general parameterized and optimized to produce visually pleasing images for the human eye. Although these pipelines can produce satisfying results with respect to their objective, there is no guarantee that this -- visually pleasing -- representation is optimal for computer vision. In fact, there are results indicating that only a handful of operations in classical ISP pipelines actually increase the performance of downstream computer vision systems \cite{buckler2017reconfiguring, olli2021end}.

Many of these handcrafted operations can be defined as learnable operations in a neural network and subsequently be optimized towards other objectives than producing visually pleasing images. Inspired by this we investigate a set of \textit{learnable} operations that are applied to the RAW image input and optimized end-to-end with respect to the downstream computer vision tasks. Inspired by the works in \cite{olli2021end,buckler2017reconfiguring, aastrom2013density, yeo2000new}, we define \textit{Learnable Gamma Correction}, \textit{Learnable Error Function}, and \textit{Learnable Yeo-Johnson}, which are described in detail below.

\begin{figure*}[t]
    \centering
    \includegraphics[width=0.95\linewidth]{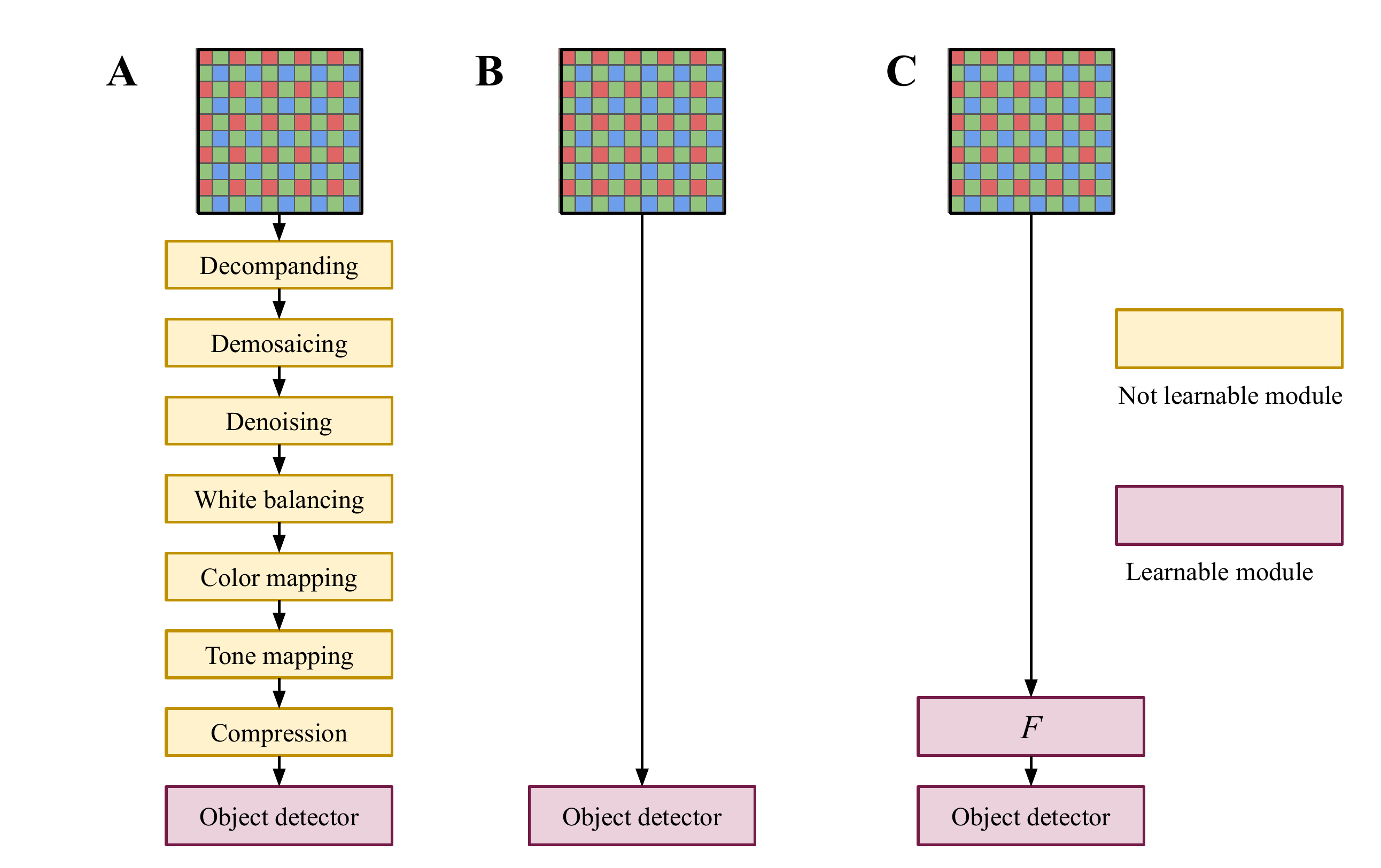} 
    \caption{Traditional (A), naïve (B), and proposed (C) detection pipelines. The traditional pipeline uses a set of common image signal processing operations, such as \textit{Demosaicing}, \textit{Denoising}, and \textit{Tonemapping}, and then feeds the object detector with the processed RGB images. The naïve pipeline feeds the RAW image directly into the detector while our proposed pipeline first feeds the RAW image through a \textit{learnable} non-linear operation, $F$, which can be viewed as being part of the end-to-end trainable object detection network.} 
    \label{fig:pipeline}
\end{figure*}

\parsection{Learnable Gamma Correction} Prior work \cite{olli2021end, buckler2017reconfiguring} has shown that the most essential operations in standard ISP pipelines are demosaicing and tone-mapping. In both works, they make use of a bilinear demosaicing algorithm together with a gamma correction method. We also implement a \textit{learnable} gamma correction defined as 
\begin{equation}
    F_\gamma(\mathbf{x}) = \mathbf{x}^{\gamma}_d \enspace, 
    \label{eq:gamma}
\end{equation}
where $\gamma\in\mathbb{R}$ is the learnable parameter that is trained jointly with the downstream network, and $\mathbf{x}_d$ is the input image $\mathbf{x}$ after bilinear demosacing. Conveniently, we can model the demosaicing operation as a 2D convolution over the entire image. By using two $3\times3$ kernels, 
\begin{equation}
    K_g =
        \begin{bmatrix}
            0.0 & 0.25 & 0.0\\
            0.25 & 1.0 & 0.25\\
            0.0 & 0.25 & 0.0
        \end{bmatrix},  \quad 
    K_{rb}=
        \begin{bmatrix}
            0.25 & 0.5 & 0.25\\
            0.5 & 1.0 & 0.5\\
            0.25 & 0.5 & 0.25\\
        \end{bmatrix} \enspace,
\end{equation} we can effectively achieve bilinear demosaicing by convolving the filters over their respective masked input. To further clarify, we convolve $K_g$ over the RAW Bayer image, where all cells that do not have the green filter are set to zero. Similarly, we convolve $R_{rb}$ over the RAW Bayer image where we only keep the red and blue cells, respectively, thus obtaining a 3-channel bilinearly interpolated RAW image. 

\parsection{Learnable Error Function} An even simpler approach is to feed the RAW input data through a single non-linear function. To this end, we adopt the Gauss error function. This function has been used in prior works to model disease cases~\cite{ciufolini2020mathematical}, as an activation function in neural networks~\cite{hendrycks2016gaussian}, and for diffusion-based image enhancement~\cite{aastrom2013density}. Formally, we define
\begin{equation}
    F_\text{erf}(\mathbf{x}) = \text{erf} \left( \frac{\mathbf{x} - \mu}{\sqrt{2} \sigma} \right) \enspace,
    \label{eq:erf}
\end{equation}
where $\mu\in\mathbb{R}$ and $\sigma\in\mathbb{R}_+$ are learnable parameters optimized jointly with the encoder and detector head parameters during training. Note that the $\text{erf}$ function saturates quickly and we found it necessary to normalize the data to be in the range of 0 to 1. 

\parsection{Learnable Yeo-Johnson transformation} A common preprocessing step in deep learning pipelines is to normalize the input data, as it has shown to improve the performance and stability of deep neural networks \cite{he2015delving, glorot2010understanding}. In object detection pipelines, this is commonly achieved by normalizing with the mean and variance of each RGB input channel across the entire dataset. While the same approach can easily be adopted to each of the colors in the Bayer pattern, this naïve approach does not yield satisfactory results. One thing to note is that work on weight initialization~\cite{he2015delving,glorot2010understanding} typically assume the input to have a standard normal distribution. We observed that the RGGB data distribution was highly non-Gaussian, motivating us to find a transformation that improves the normality of the data.
 
Yeo and Johnson proposed a new family of power transformations that aims to improve the symmetry and normality of the transformed data \cite{yeo2000new}. These transformations are parameterized by $\lambda$, which is usually optimized offline by maximizing the log-likelihood between the input data and a Gaussian distribution. However, analogously to the ISP operations that should be optimized towards the end task, we can optimize the Yeo-Johnson transformation with respect to the end goal, rather than towards a Gaussian distribution.  Inspired by this, we define the \textit{Learnable Yeo-Johnson} transformation as a point-wise non-linear operation
\begin{equation}
    F_\text{YJ}(\mathbf{x}) = \frac{(\mathbf{x}+1)^{\lambda} - 1}{\lambda} \enspace,
    \label{eq:yeojohnson}
\end{equation} 
where $\lambda\in\mathbb{R}_+$ is the learnable parameter. 

\subsection{Our Raw Object Detector}
Given RAW RGGB images, we downsample as described in Section~\ref{section:downsampling} to obtain $\mathbf{x}$. Then, we apply one of the learnable ISP operations, $F$, as described in \eqref{eq:gamma}, \eqref{eq:erf}, or \eqref{eq:yeojohnson}. Finally, we apply the object detector, $D$,
\begin{align}
\mathcal{O} = D(F(\mathbf{x}))\enspace,
\end{align}
giving us a set of predicted objects $\mathcal{O}$. We train $F$ and $D$ jointly.

\section{Experiments}
In this section, we introduce the dataset on which we evaluate the different methods (Section \ref{sec:dataset}), along with some of the prominent implementation details (Section \ref{sec:implementation}) used during training and evaluation. Next, we present the results, both quantitative (Section \ref{sec:quantative-results}) and qualitative (Section \ref{sec:qualatitive-results}) for all the learnable operations proposed in Section \ref{sec:learnableisp}. Lastly, we present how the learnable parameters in each of the proposed operations evolve during training in Section \ref{sec:paramter-evolution}.

\subsection{Dataset}
\label{sec:dataset}

To evaluate our learnable operations, we make use of the PASCALRAW dataset \cite{omid2014pascalraw}. This dataset contains 4259 high-resolution ($6034 \times 4012$) RAW 12bit RGGB images, all captured with a Nikon D3200 DSLR camera during daylight conditions in Palo Alto and San Francisco. We downsample all RAW images to a resolution more compatible with standard object detection pipelines ($1206 \times 802$) according to the Bayer-pattern-preserving downsampling described in Section~\ref{section:downsampling}. Note that we crop away the last four rows and two columns ($0.1\%$ of the image) to obtain an integer downsampling factor. Subsequently, we generate the corresponding RGB images (used by the RGB Baseline) from the downsampled RAW images using a standard ISP pipeline implemented in the RAW image processing library RawPy \cite{rawpy2022}. For each image, the authors provide dense annotations in the form of class-bounding-box-pairs for three different classes: pedestrian, car, and bicycle. In total, the dataset contains 6550 annotated instances, divided into 4077 pedestrians, 1765 cars, and 708 bicycles.

\subsection{Implementation details}
\label{sec:implementation}
We use a standard object detection pipeline, namely a Faster-RCNN \cite{ren2015faster}, with a Feature Pyramid Network \cite{lin2017feature}, and a ResNet-50 \cite{he2016deep} backbone. All models were implemented, trained, and evaluated in the Detectron2  framework \cite{wu2019detectron2}. We use a batch size of $B = 16$, a learning rate of $l_r = 3\cdot10^{-4}$, a learning-rate scheduler with 5000 warm-up iterations, and a learning-rate drop by a factor $\alpha = 0.1$ after 100k iterations. We train for 150k iterations using an SGD optimizer. The learnable parameters in the ISP pipeline, $\lambda$, $\gamma$, $\mu$, and $\sigma$, were initialized (when used) to $0.35$, $1.0$, $1.0$, and $1.0$ respectively.

\subsection{Quantitative Results}
\label{sec:quantative-results}
In Table \ref{experiments:pascal-raw-results} we present the results when training and evaluating our different learnable functions on the PASCALRAW dataset. The results are presented in terms of \emph{mean average precision} (AP), following the COCO detection benchmark~\cite{lin2014microsoft}. We also provide average precision for different IoU-thresholds (AP$_{50}$ and AP$_{75}$) and AP for each class. We report the mean and standard deviation over three separate runs.
\begin{table}[t]
    \centering
    \caption{Object detection results on the PASCALRAW dataset. The results are presented in terms of AP (higher is better) and we report the mean and standard deviation over 3 separate runs.}
    \label{experiments:pascal-raw-results}
    \resizebox{\textwidth}{!}{%
        \begin{tabular}{lcccccc}
            \hline
            Components                 & $\text{AP}$                        & $\text{AP}_{50}$                   & $\text{AP}_{75}$                   & $\text{AP}_{car}$                  & $\text{AP}_{ped}$                  & $\text{AP}_{bic}$     \\
            \hline
            RGB Baseline               & 50.5 $\pm$ 0.5 \hspace{5pt} & 84.8 $\pm$ 0.3 \hspace{5pt} & 55.2 $\pm$ 1.6 \hspace{5pt} & 61.8 $\pm$ 0.1 \hspace{5pt} & 48.5 $\pm$ 0.7 \hspace{5pt} & 41.4 $\pm$ 0.8 \\
            RAW RGGB Baseline          & 31.3 $\pm$ 1.2 \hspace{5pt} & 64.7 $\pm$ 1.6 \hspace{5pt} & 25.2 $\pm$ 2.0 \hspace{5pt} & 42.4 $\pm$ 1.8 \hspace{5pt} & 30.5 $\pm$ 0.5 \hspace{5pt} & 20.9 $\pm$ 1.5 \\
            RAW + Learnable Gamma          & 51.4 $\pm$ 0.3 \hspace{5pt} & 85.8 $\pm$ 0.6 \hspace{5pt} & 56.3 $\pm$ 0.7 \hspace{5pt} & 62.5 $\pm$ 0.4 \hspace{5pt} & 49.0 $\pm$ 0.2 \hspace{5pt} & 42.7 $\pm$ 1.1 \\
            RAW + Learnable Error Function & 49.3 $\pm$ 0.2 \hspace{5pt} & 84.0 $\pm$ 0.4 \hspace{5pt} & 52.8 $\pm$ 0.5 \hspace{5pt} & 60.1 $\pm$ 0.6 \hspace{5pt} & 46.3 $\pm$ 0.5 \hspace{5pt} & 41.3 $\pm$ 0.8 \\
            RAW + Learnable Yeo-Johnson   & \textbf{52.6 $\pm$ 0.4} \hspace{5pt} & \textbf{86.7 $\pm$ 0.3} \hspace{5pt} & \textbf{57.9 $\pm$ 0.6} \hspace{5pt} & \textbf{63.6 $\pm$ 0.5} \hspace{5pt} & \textbf{49.9 $\pm$ 0.4} \hspace{5pt} &\textbf{ 44.2 $\pm$ 0.6} \\
            \hline
        \end{tabular}
    }
\end{table}

From the results in Table \ref{experiments:pascal-raw-results}, we can conclude that simply feeding the RAW RGGB image (i.e., removing all ISP operations) into a standard object detection network, corresponding to the RAW RGGB Baseline in Figure \ref{fig:pipeline}(B), performs substantially worse than the traditional RGB Baseline in Figure \ref{fig:pipeline}(A). Further, we can corroborate the results of \cite{olli2021end, buckler2017reconfiguring} and observe that the method RAW + \emph{Learnable Gamma}, which comprises the two operations \emph{demosaicing} and \emph{gamma correction}, by a slight margin surpasses the performance of the RGB Baseline. Lastly, we also observe that our method RAW +\textit{Learnable Yeo-Johnson} in Figure~\ref{fig:pipeline}(C) outperforms all other methods by a statistically significant margin.

\subsection{Qualitative Results}
\label{sec:qualatitive-results}
From Table \ref{experiments:pascal-raw-results} it is evident that our \textit{Learnable Yeo-Johnson} operation outperforms the RGB baseline. We hypothesize that this is partly because our learnable ISP can better handle poor (low) light conditions. In Figure \ref{fig:samples-raw-better}, we present three examples from the PASCALRAW test set that further support this hypothesis. Our RAW image pipeline can more accurately detect objects in the darker parts of the images, whereas the RGB Baseline fails in the same situations.

\subsection{Parameter Evolution}
\label{sec:paramter-evolution}
\begin{figure*}[thb]
    \centering
    \begin{tabular}{cc}
         \includegraphics[width=0.46\linewidth]{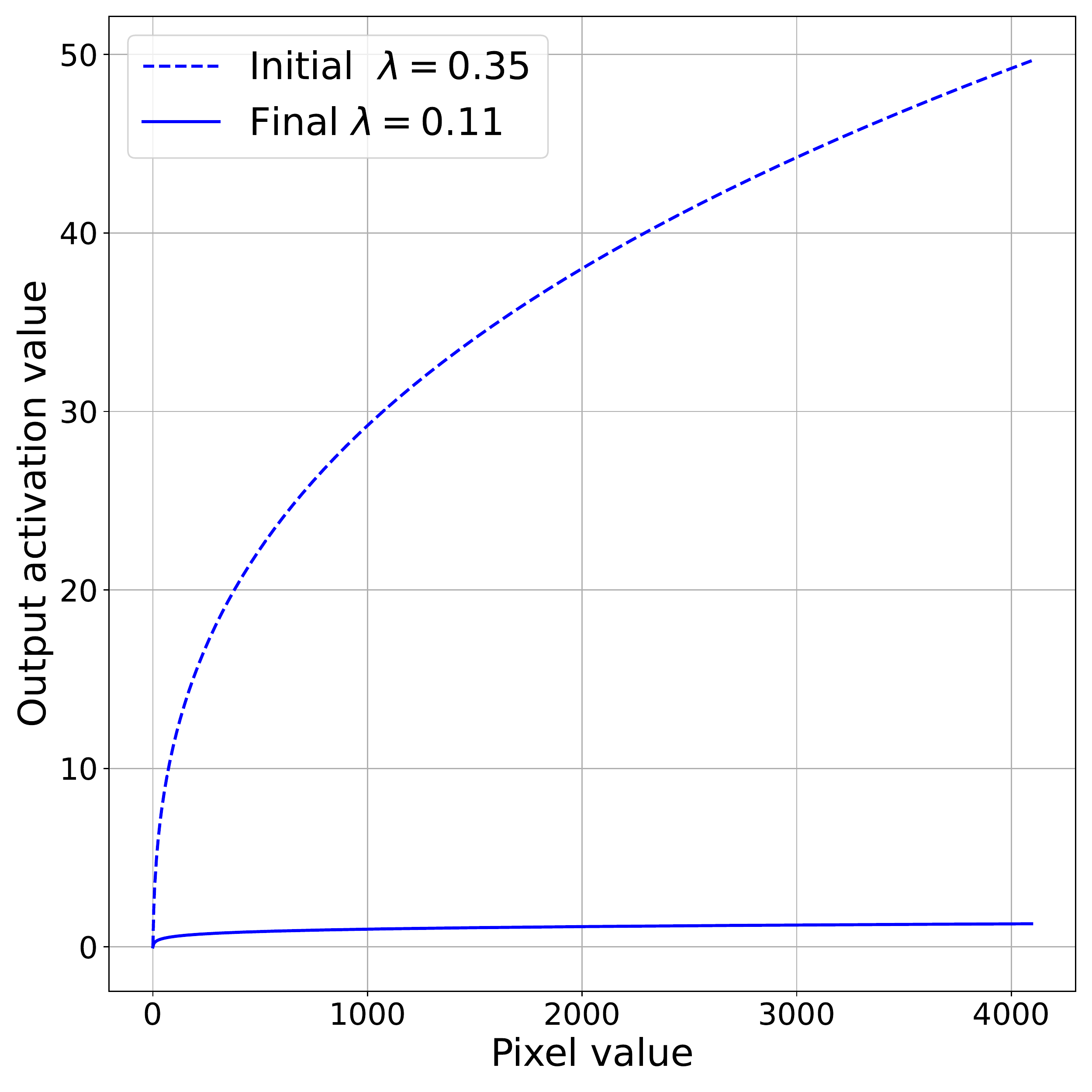} & \includegraphics[width=0.46\linewidth]{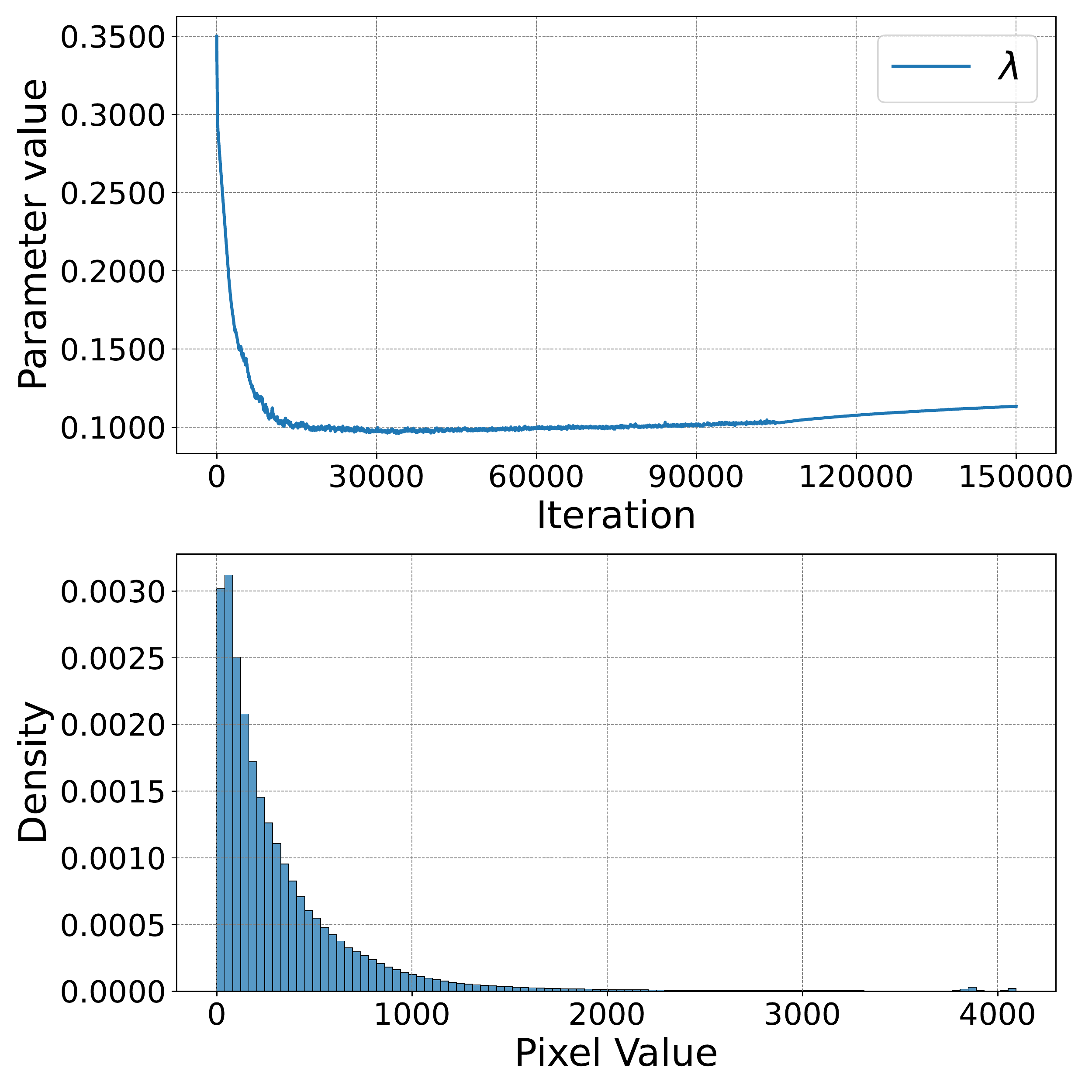}
    \end{tabular}
    
    \caption{Evolution of the learnable parameter $\lambda$ during the entire training (top-right), the distribution of the RAW pixel values in PASCAL RAW (bottom-right), and the functional form  -- before and after training -- of the \textit{Learnable Yeo-Johnson} operation (left). In the left plot, the output activation values are shown across the full input range $[0, 2^{12}-1$].}
    \label{fig:parameter-evolution}
\end{figure*}

To further analyze the behavior of our \textit{Learnable Yeo-Johnson} operation, we show the evolution of its trainable parameter, $\lambda$, along with the functional form of the operation, in Figure \ref{fig:parameter-evolution}. We observe that the training converges to a relatively low value of $\lambda$, which, as can be seen from the functional form of the operation, implies that low-valued/dark pixels are better differentiated than high-valued/bright pixels. This characteristic suggests that the RAW object detector is able to better distinguish features in low-light regions of the image, compared to the RGB detector, thus achieving better detection performance.

\section{Conclusion}
Motivated by the observation that camera ISP pipelines are typically optimized towards producing visually pleasing images for the human eye, we have in this work experimented with object detection on RAW images. While naïvely feeding RAW images directly into the object detection backbone led to poor performance, we proposed three simple, learnable operations that all led to good performance. Two of these operators, the \textit{Learnable Gamma} and \textit{Learnable Yeo-Johnson}, led to superior performance compared to the RGB baseline detector. Based on qualitative comparison, the RAW detector performs better in low-light conditions compared to the RGB detector.

\subsubsection{Acknowledgements} This work was partially supported by the Wallenberg AI, Autonomous Systems and Software Program (WASP) funded by the Knut and Alice Wallenberg Foundation.

\bibliographystyle{splncs04}
\bibliography{paper}

\begin{thebibliography}{10}
\providecommand{\url}[1]{\texttt{#1}}
\providecommand{\urlprefix}{URL }
\providecommand{\doi}[1]{https://doi.org/#1}

\bibitem{aastrom2013density}
{\AA}str{\"o}m, F., Zografos, V., Felsberg, M.: Density driven diffusion. In:
  Scandinavian Conference on Image Analysis. pp. 718--730. Springer (2013)

\bibitem{bayer1976color}
Bayer, B.E.: Color imaging array. United States Patent 3,971,065  (1976)

\bibitem{buades2005non}
Buades, A., Coll, B., Morel, J.M.: A non-local algorithm for image denoising.
  In: 2005 IEEE computer society conference on computer vision and pattern
  recognition (CVPR'05). vol.~2, pp. 60--65. Ieee (2005)

\bibitem{buckler2017reconfiguring}
Buckler, M., Jayasuriya, S., Sampson, A.: Reconfiguring the imaging pipeline
  for computer vision. In: Proceedings of the IEEE International Conference on
  Computer Vision. pp. 975--984 (2017)

\bibitem{carion2020end}
Carion, N., Massa, F., Synnaeve, G., Usunier, N., Kirillov, A., Zagoruyko, S.:
  End-to-end object detection with transformers. In: European conference on
  computer vision. pp. 213--229. Springer (2020)

\bibitem{ciufolini2020mathematical}
Ciufolini, I., Paolozzi, A.: Mathematical prediction of the time evolution of
  the covid-19 pandemic in italy by a gauss error function and monte carlo
  simulations. The European Physical Journal Plus  \textbf{135}(4), ~355 (2020)

\bibitem{condat2010simple}
Condat, L.: A simple, fast and efficient approach to denoisaicking: Joint
  demosaicking and denoising. In: 2010 IEEE International Conference on Image
  Processing. pp. 905--908. IEEE (2010)

\bibitem{dai2020awnet}
Dai, L., Liu, X., Li, C., Chen, J.: Awnet: Attentive wavelet network for image
  isp. In: European Conference on Computer Vision. pp. 185--201. Springer
  (2020)

\bibitem{dubois2006filter}
Dubois, E.: Filter design for adaptive frequency-domain bayer demosaicking. In:
  2006 International Conference on Image Processing. pp. 2705--2708. IEEE
  (2006)

\bibitem{foi2008practical}
Foi, A., Trimeche, M., Katkovnik, V., Egiazarian, K.: Practical
  poissonian-gaussian noise modeling and fitting for single-image raw-data.
  IEEE Transactions on Image Processing  \textbf{17}(10),  1737--1754 (2008)

\bibitem{girshick2014rich}
Girshick, R., Donahue, J., Darrell, T., Malik, J.: Rich feature hierarchies for
  accurate object detection and semantic segmentation. In: Proceedings of the
  IEEE conference on computer vision and pattern recognition. pp. 580--587
  (2014)

\bibitem{glorot2010understanding}
Glorot, X., Bengio, Y.: Understanding the difficulty of training deep
  feedforward neural networks. In: Proceedings of the thirteenth international
  conference on artificial intelligence and statistics. pp. 249--256. JMLR
  Workshop and Conference Proceedings (2010)

\bibitem{he2015delving}
He, K., Zhang, X., Ren, S., Sun, J.: Delving deep into rectifiers: Surpassing
  human-level performance on imagenet classification. In: Proceedings of the
  IEEE international conference on computer vision. pp. 1026--1034 (2015)

\bibitem{he2016deep}
He, K., Zhang, X., Ren, S., Sun, J.: Deep residual learning for image
  recognition. In: Proceedings of the IEEE conference on computer vision and
  pattern recognition. pp. 770--778 (2016)

\bibitem{hendrycks2016gaussian}
Hendrycks, D., Gimpel, K.: Gaussian error linear units (gelus). arXiv preprint
  arXiv:1606.08415  (2016)

\bibitem{hirakawa2005adaptive}
Hirakawa, K., Parks, T.W.: Adaptive homogeneity-directed demosaicing algorithm.
  Ieee transactions on image processing  \textbf{14}(3),  360--369 (2005)

\bibitem{hong2021crafting}
Hong, Y., Wei, K., Chen, L., Fu, Y.: Crafting object detection in very low
  light. In: BMVC. vol.~1, p.~3 (2021)

\bibitem{hp2020autoencoder}
HP, A.W., Prasetyo, H., Guo, J.M.: Autoencoder-based image companding. In: 2020
  IEEE International Conference on Consumer Electronics-Taiwan (ICCE-Taiwan).
  pp.~1--2. IEEE (2020)

\bibitem{ignatov2020replacing}
Ignatov, A., Van~Gool, L., Timofte, R.: Replacing mobile camera isp with a
  single deep learning model. In: Proceedings of the IEEE/CVF Conference on
  Computer Vision and Pattern Recognition Workshops. pp. 536--537 (2020)

\bibitem{krawczyk2005lightness}
Krawczyk, G., Myszkowski, K., Seidel, H.P.: Lightness perception in tone
  reproduction for high dynamic range images. In: Computer Graphics Forum.
  vol.~24, pp. 635--646. Amsterdam: North Holland, 1982- (2005)

\bibitem{kriesel2014traue}
Kriesel, D.: Traue keinem scan, den du nicht selbst gef{\"a}lscht hast.
  Mitteilungen der Deutschen Mathematiker-Vereinigung  \textbf{22}(1),  30--34
  (2014)

\bibitem{langseth2014evaluation}
Langseth, R., Gaddam, V.R., Stensland, H.K., Griwodz, C., Halvorsen, P.: An
  evaluation of debayering algorithms on gpu for real-time panoramic video
  recording. In: 2014 IEEE International Symposium on Multimedia. pp. 110--115.
  IEEE (2014)

\bibitem{li2008image}
Li, X., Gunturk, B., Zhang, L.: Image demosaicing: A systematic survey. In:
  Visual Communications and Image Processing 2008. vol.~6822, pp. 489--503.
  SPIE (2008)

\bibitem{lin2017feature}
Lin, T.Y., Doll{\'a}r, P., Girshick, R., He, K., Hariharan, B., Belongie, S.:
  Feature pyramid networks for object detection. In: Proceedings of the IEEE
  conference on computer vision and pattern recognition. pp. 2117--2125 (2017)

\bibitem{lin2017focal}
Lin, T.Y., Goyal, P., Girshick, R., He, K., Doll{\'a}r, P.: Focal loss for
  dense object detection. In: Proceedings of the IEEE international conference
  on computer vision. pp. 2980--2988 (2017)

\bibitem{lin2014microsoft}
Lin, T.Y., Maire, M., Belongie, S., Hays, J., Perona, P., Ramanan, D.,
  Doll{\'a}r, P., Zitnick, C.L.: Microsoft coco: Common objects in context. In:
  European conference on computer vision. pp. 740--755. Springer (2014)

\bibitem{liu2021swin}
Liu, Z., Lin, Y., Cao, Y., Hu, H., Wei, Y., Zhang, Z., Lin, S., Guo, B.: Swin
  transformer: Hierarchical vision transformer using shifted windows. In:
  Proceedings of the IEEE/CVF International Conference on Computer Vision. pp.
  10012--10022 (2021)

\bibitem{malvar2004high}
Malvar, H.S., He, L.w., Cutler, R.: High-quality linear interpolation for
  demosaicing of bayer-patterned color images. In: 2004 IEEE International
  Conference on Acoustics, Speech, and Signal Processing. vol.~3, pp. iii--485.
  IEEE (2004)

\bibitem{meng2021conditional}
Meng, D., Chen, X., Fan, Z., Zeng, G., Li, H., Yuan, Y., Sun, L., Wang, J.:
  Conditional detr for fast training convergence. In: Proceedings of the
  IEEE/CVF International Conference on Computer Vision. pp. 3651--3660 (2021)

\bibitem{morawski2022genisp}
Morawski, I., Chen, Y.A., Lin, Y.S., Dangi, S., He, K., Hsu, W.H.: Genisp:
  Neural isp for low-light machine cognition. In: Proceedings of the IEEE/CVF
  Conference on Computer Vision and Pattern Recognition. pp. 630--639 (2022)

\bibitem{mujtaba2022efficient}
Mujtaba, N., Khan, I.R., Khan, N.A., Altaf, M.A.B.: Efficient flicker-free tone
  mapping of hdr videos. In: 2022 IEEE 24th International Workshop on
  Multimedia Signal Processing (MMSP). pp. 01--06. IEEE (2022)

\bibitem{olli2021end}
Olli~Blom, M., Johansen, T.: End-to-end object detection on raw camera data
  (2021)

\bibitem{omid2014pascalraw}
Omid-Zohoor, A., Ta, D., Murmann, B.: Pascalraw: raw image database for object
  detection (2014)

\bibitem{poynton2012digital}
Poynton, C.: Digital video and HD: Algorithms and Interfaces. Elsevier (2012)

\bibitem{redmon2018yolov3}
Redmon, J., Farhadi, A.: Yolov3: An incremental improvement. arXiv preprint
  arXiv:1804.02767  (2018)

\bibitem{reinhard2002photographic}
Reinhard, E., Stark, M., Shirley, P., Ferwerda, J.: Photographic tone
  reproduction for digital images. In: Proceedings of the 29th annual
  conference on Computer graphics and interactive techniques. pp. 267--276
  (2002)

\bibitem{ren2015faster}
Ren, S., He, K., Girshick, R., Sun, J.: Faster r-cnn: Towards real-time object
  detection with region proposal networks. Advances in neural information
  processing systems  \textbf{28} (2015)

\bibitem{rawpy2022}
Riechert, M.: Rawpy. \url{https://github.com/letmaik/rawpy} (2022)

\bibitem{shekhar2022transform}
Shekhar~Tripathi, A., Danelljan, M., Shukla, S., Timofte, R., Van~Gool, L.:
  Transform your smartphone into a dslr camera: Learning the isp in the wild.
  In: European Conference on Computer Vision. pp. 625--641. Springer (2022)

\bibitem{suma2016evaluation}
Suma, R., Stavropoulou, G., Stathopoulou, E.K., Van~Gool, L., Georgopoulos, A.,
  Chalmers, A.: Evaluation of the effectiveness of hdr tone-mapping operators
  for photogrammetric applications. Virtual Archaeology Review  \textbf{7}(15),
   54--66 (2016)

\bibitem{sun2021rethinking}
Sun, Z., Cao, S., Yang, Y., Kitani, K.M.: Rethinking transformer-based set
  prediction for object detection. In: Proceedings of the IEEE/CVF
  international conference on computer vision. pp. 3611--3620 (2021)

\bibitem{tian2019fcos}
Tian, Z., Shen, C., Chen, H., He, T.: Fcos: Fully convolutional one-stage
  object detection. In: Proceedings of the IEEE/CVF international conference on
  computer vision. pp. 9627--9636 (2019)

\bibitem{wang2022anchor}
Wang, Y., Zhang, X., Yang, T., Sun, J.: Anchor detr: Query design for
  transformer-based detector. In: Proceedings of the AAAI conference on
  artificial intelligence. vol.~36, pp. 2567--2575 (2022)

\bibitem{wu2019detectron2}
Wu, Y., Kirillov, A., Massa, F., Lo, W.Y., Girshick, R.: Detectron2.
  \url{https://github.com/facebookresearch/detectron2} (2019)

\bibitem{yeo2000new}
Yeo, I.K., Johnson, R.A.: A new family of power transformations to improve
  normality or symmetry. Biometrika  \textbf{87}(4),  954--959 (2000)

\bibitem{yoshimura2022dynamicisp}
Yoshimura, M., Otsuka, J., Irie, A., Ohashi, T.: Dynamicisp: Dynamically
  controlled image signal processor for image recognition. arXiv preprint
  arXiv:2211.01146  (2022)

\bibitem{yoshimura2022rawgment}
Yoshimura, M., Otsuka, J., Irie, A., Ohashi, T.: Rawgment: Noise-accounted raw
  augmentation enables recognition in a wide variety of environments. arXiv
  preprint arXiv:2210.16046  (2022)

\bibitem{zhang2022dino}
Zhang, H., Li, F., Liu, S., Zhang, L., Su, H., Zhu, J., Ni, L.M., Shum, H.Y.:
  Dino: Detr with improved denoising anchor boxes for end-to-end object
  detection. arXiv preprint arXiv:2203.03605  (2022)

\bibitem{zhang2021raw}
Zhang, X., Zhang, L., Lou, X.: A raw image-based end-to-end object detection
  accelerator using hog features. IEEE Transactions on Circuits and Systems I:
  Regular Papers  \textbf{69}(1),  322--333 (2021)

\bibitem{zhang2021learning}
Zhang, Z., Wang, H., Liu, M., Wang, R., Zhang, J., Zuo, W.: Learning
  raw-to-srgb mappings with inaccurately aligned supervision. In: Proceedings
  of the IEEE/CVF International Conference on Computer Vision. pp. 4348--4358
  (2021)

\bibitem{zhou2019objects}
Zhou, X., Wang, D., Kr{\"a}henb{\"u}hl, P.: Objects as points. arXiv preprint
  arXiv:1904.07850  (2019)

\bibitem{zhu2020deformable}
Zhu, X., Su, W., Lu, L., Li, B., Wang, X., Dai, J.: Deformable detr: Deformable
  transformers for end-to-end object detection. arXiv preprint arXiv:2010.04159
   (2020)

\end{thebibliography}

\end{document}